\documentclass[letterpaper, 10 pt, conference]{ieeeconf}

\IEEEoverridecommandlockouts                            
\usepackage{color}
\usepackage{array}
\usepackage{tabularx}
\usepackage{amsmath,amsfonts}
\usepackage{fancyhdr}
\usepackage{amssymb}
\usepackage{type1cm}
\usepackage{url}
\usepackage{caption}
\usepackage[pdftex]{graphicx}
\usepackage{bm}
\usepackage{graphicx}
\usepackage{multirow}  
\usepackage{ulem}
\usepackage{cite}

\usepackage{stfloats}
\usepackage{textcomp}
\usepackage{verbatim}

\usepackage{ifthen}
\usepackage{animate}
\usepackage{bm}
\usepackage[bookmarks=true]{hyperref}
\usepackage{booktabs}
\newcommand{\thickhline}{\noalign{\hrule height 1.2pt}}

\usepackage{algorithm}
\usepackage{algpseudocode}

\usepackage{xcolor}

\usepackage{listings}
\usepackage{xcolor}
\begin{document}

\title{\LARGE \bf
Embodied Tree of Thoughts: Deliberate Manipulation Planning with Embodied World Model
}

\author{
    Wenjiang Xu$^{1,5}$,
    Cindy Wang$^{2}$,
    Rui Fang$^{2}$,
    Mingkang Zhang$^{2}$,
    Lusong Li$^{3}$,
    Jing Xu $^2$,
    Jiayuan Gu$^{4}$,\\
    Zecui Zeng$^{3}$$^{\dagger}$,
    Rui Chen$^{2}$$^{\dagger}$
    \vspace{0.5mm} \\
    $^1$University of Chinese Academy of Sciences (UCAS) \quad 
    $^2$Tsinghua University\quad 
    $^3$JD Explore Academy  \\
    $^4$ShanghaiTech University \quad 
    $^5$Nanjing University  \\
    \vspace{2mm}
    \url{https://embodied-tree-of-thoughts.github.io/}  \vspace{-4mm}
}
\twocolumn[{
    \renewcommand\twocolumn[1][]{#1}
    \maketitle
    \begin{center}
    \captionsetup{type=figure}
    \setcounter{figure}{0}
    \includegraphics[trim=0.4ex 0 0 0, clip, width=1.0\textwidth]{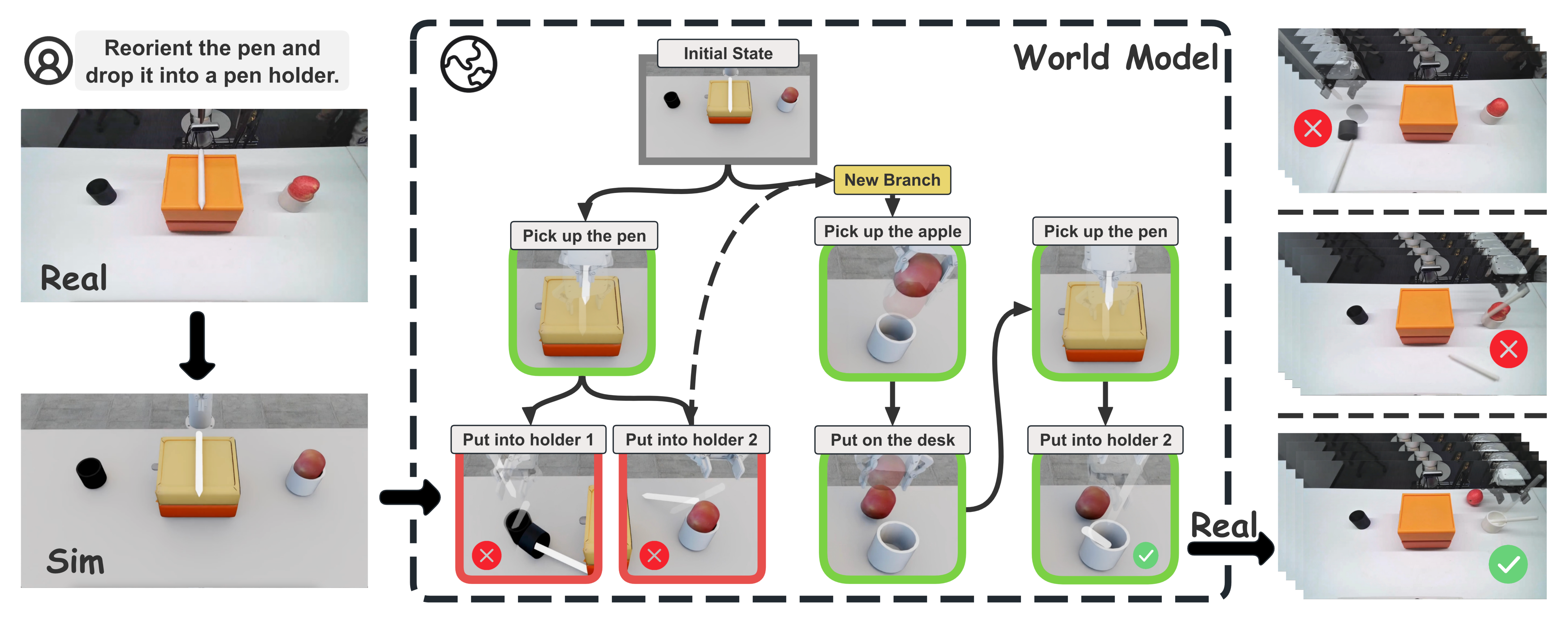}
    \caption{We propose \textbf{Embodied Tree of Thoughts (EToT)}, a Real2Sim2Real manipulation planning framework grounded in an embodied world model. EToT integrates two synergistic mechanisms—\textit{Priori Branching} (solid arrows), which enumerates candidate plan branches, and \textit{Reflective Branching} (dashed arrows), which refines the tree based on simulated execution outcomes—to iteratively expand and search the planning tree before executing the feasible plan in the real world.} 
    \label{fig:teaser}
    \end{center}
    \vspace{2.5mm}
}]

\begin{abstract}
World models have emerged as a pivotal component in robot manipulation planning, enabling agents to predict future environmental states and reason about the consequences of actions before execution. While video-generation models are increasingly adopted, they often lack rigorous physical grounding, leading to hallucinations and a failure to maintain consistency in long-horizon physical constraints. To address these limitations, we propose Embodied Tree of Thoughts (EToT), a novel Real2Sim2Real planning framework that leverages a physics-based interactive digital twin as an embodied world model. EToT formulates manipulation planning as a tree search expanded through two synergistic mechanisms: (1) Priori Branching, which generates diverse candidate execution paths based on semantic and spatial analysis; and (2) Reflective Branching, which utilizes VLMs to diagnose execution failures within the simulator and iteratively refine the planning tree with corrective actions. By grounding high-level reasoning in a physics simulator, our framework ensures that generated plans adhere to rigid-body dynamics and collision constraints. We validate EToT on a suite of short- and long-horizon manipulation tasks, where it consistently outperforms baselines by effectively predicting physical dynamics and adapting to potential failures.

\end{abstract}

\footnotetext{$^{\dagger}$ Corresponding author.}

\section{INTRODUCTION}

Developing a general-purpose robotic system capable of accomplishing complex manipulation tasks in open-world environments remains a fundamental challenge~\cite{doi:10.1126/scirobotics.abd9461}. Such systems must bridge high-level semantic understanding with low-level physical execution. Recent advances in Vision-Language models (VLMs)~\cite{achiam2023gpt, liu2023visual} have enabled robots to interpret natural language instructions and generate high-level task plans~\cite{hu2023look, huang2022innermonologueembodiedreasoning, shah2025bumble}. However, these approaches primarily operate on static scene representations and lack the physical intuition required to predict the dynamic evolution of the environment under long-horizon action sequences.

To address this limitation, world models that predict future states conditioned on robot actions have attracted increasing attention. A line of recent work adopts video generation models as forward predictors of future scenes for grounded planning~\cite{feng2025reflectiveplanningvisionlanguagemodels}. While effective for short-term prediction, such pixel-space models lack explicit physical grounding and struggle to capture the cumulative effects of contact-rich interactions, often producing physically inconsistent ``hallucinations'' over long horizons~\cite{zhang2025worldinworld}. As a result, their applicability is typically restricted to short-horizon action prediction where physical consistency is less critical.

An alternative paradigm is Real2Sim, which reconstructs real-world scenes within physics simulators and leverages the simulator as a physically grounded world model~\cite{ning2025prompting}. Recent breakthroughs in 3D AIGC~\cite{liu2024one, sam3dteam2025sam3d3dfyimages} and the maturation of high-fidelity simulation platforms~\cite{li2024behavior1k, taomaniskill3} have significantly improved the feasibility of this approach. Unlike video-based predictors, simulator-based world models enforce explicit physical laws, enabling consistent multi-step dynamics and reliable modeling of contact interactions. Moreover, simulators provide direct access to latent physical properties such as mass, friction, and joint constraints, which are essential for accurate long-horizon planning.

In this work, we introduce \textit{Embodied Tree of Thoughts (EToT)}, a planning framework that grounds VLM-based reasoning in a physics-based embodied world model. In contrast to video-generation approaches, EToT employs a physics simulator to ensure that all predicted outcomes strictly adhere to rigid-body dynamics and collision constraints. Furthermore, because real-world manipulation tasks often exhibit multiple alternative action choices and long-range causal dependencies, we formulate task planning as a tree-structured search process that provides sufficient breadth and depth to explore feasible solutions.

As illustrated in Fig.~\ref{fig:teaser}, we reconstruct the real-world scene as an interactive digital twin within a physics simulator. This physics-grounded twin enables the planner to simulate the outcomes of VLM-generated actions prior to physical execution and perform visual failure analysis. Through \textit{Priori Branching}, the planner generates diverse candidate action sequences that form the initial planning tree. When a node fails in simulation, the corresponding simulated observations are fed back to the VLM, which performs \textit{Reflective Branching} to analyze the failure cause and generate revised branches based on the original plan. Through this iterative loop of simulation, visual diagnosis, and tree expansion, EToT progressively uncovers physically validated plans for complex, long-horizon real-world tasks.

To systematically evaluate the proposed framework, we construct a suite of real-world tabletop manipulation tasks ranging from short-horizon interactions to multi-stage rearrangement problems. Experimental results demonstrate that by explicitly reasoning over an embodied world model, evaluating the physical feasibility of candidate actions, and iteratively refining the planning tree, EToT significantly outperforms existing baselines. These findings highlight the importance of predicting long-term physical consequences, identifying latent failure factors, and adaptively refining plans for achieving reliable and robust robotic manipulation.

\section{RELATED WORK}

\subsection{World Models for Manipulation Planning}
\label{sec:world_model} 
A world model aims to predict the future evolution of the environment under candidate robot actions~\cite{li2025comprehensive}. Prior approaches to world modeling for manipulation planning span multiple paradigms, including textual reasoning, video prediction, 3D generative modeling, and physics-based simulation. Early reflective methods~\cite{lan2025experiencebestteachergrounding, liu2023reflect} function as textual world models that anticipate outcomes based on prior experience. However, these methods rely on coarse spatial abstractions and lack physically grounded forecasting.

More recent works~\cite{zhao2024vlmpc, feng2025reflectiveplanningvisionlanguagemodels} employ video generation models to hallucinate future scene states. Similarly, 3D flow- or Gaussian-based representations~\cite{zhi20253dflowactionlearningcrossembodimentmanipulation, lu2024manigaussian} model environmental dynamics by learning pixel-wise or point-wise deformations over time. Despite their expressiveness, these approaches typically lack explicit physical constraints and struggle to maintain long-range causal consistency, limiting their effectiveness in long-horizon, multi-stage reasoning and in satisfying zero-shot logical constraints (e.g., detecting occluded geometric interference)~\cite{liang2025videogeneratorsrobotpolicies, zhang2025worldinworld}.

In contrast, physics-based simulators provide explicit and generalizable access to rigid-body dynamics, contact interactions, and gravity, enabling high-fidelity prediction. Advanced simulators such as OmniGibson further support rich and structured embodied environments~\cite{li2024behavior1k}. Closest to our work, PWTF~\cite{ning2025prompting} employs an interactive digital twin for model predictive control by sampling low-level actions, rendering predicted outcomes, and evaluating them with a VLM. However, its use of the world model is confined to low-level control, while high-level planning remains restricted to a single, fixed task decomposition. Without the ability to critique or revise high-level plans using world-model feedback, any initial decomposition error irreversibly leads to task failure.

In contrast, our approach incorporates two complementary mechanisms for high-level planning. \textit{Priori Branching} generates multiple candidate plan branches for evaluation by the world model, thereby avoiding the single-path limitation of prior work. When failures arise due to spatial or physical constraints, \textit{Reflective Branching} analyzes realistic simulated rollouts to diagnose errors and synthesize revised branches.

\subsection{VLMs for Manipulation Planning}

\begin{figure*}[!t]
    \centering
    \includegraphics[width=0.95\textwidth]{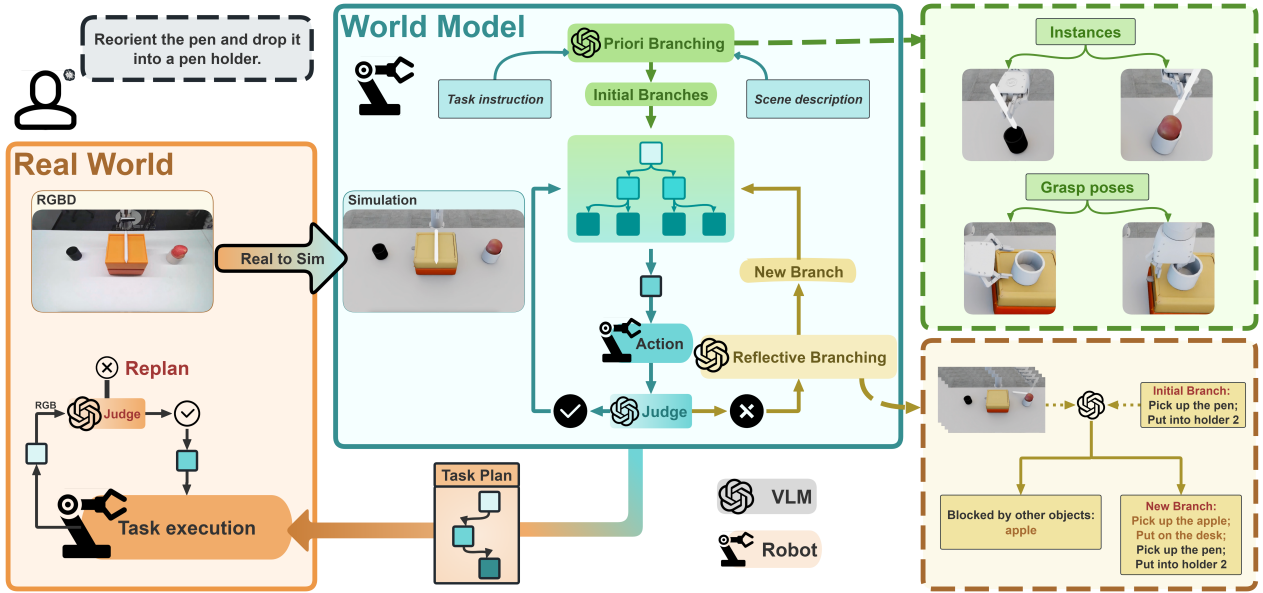}
    \caption{Overview of the \textbf{Embodied Tree of Thoughts (EToT)} framework.
Given a task instruction, the system first reconstructs the real scene into an interactive 3D digital twin (Sec.~\ref{sec:Real2Sim}). It then constructs a world-model-grounded planning tree through \textit{Priori Branching} and \textit{Reflective Branching} (Sec.~\ref{sec:plan-tree}). Priori Branching proposes initial candidate branches, while Reflective Branching analyzes simulated execution failures to expand the tree with revised branches. Through iterative searching and expansion of the planning tree, the system identifies a feasible plan, which is finally executed on the real robot in a closed-loop manner with visual feedback and re-planning (Sec.~\ref{sec:closed-loop}).
}
    \label{fig:etot}
    \vspace{-3mm}
\end{figure*}

Recent advances in LLMs~\cite{brown2020language, touvron2023llama} and VLMs~\cite{achiam2023gpt, liu2023visual} have enabled robots to interpret complex visual scenes and generate high-level task plans from natural language instructions~\cite{hu2023look, huang2022innermonologueembodiedreasoning, shah2025bumble}. However, these plans are often expressed in abstract semantic terms and lack the spatial precision required for direct execution.

To bridge this gap, several methods~\cite{huang2024copa, huang2024rekep, ji2025robobrain} integrate vision models to predict pixel-level keypoints or directional constraints for grounding high-level reasoning in executable geometry. OmniManip~\cite{pan2025omnimanip} further reconstructs 3D object models to infer more reliable manipulation strategies. Nevertheless, these approaches rely primarily on static geometric observations and lack a physics engine for predicting the dynamic evolution of the environment or the long-term consequences of contact-rich interactions. 

In this work, we directly address these limitations by integrating a simulation-based world model that enables VLMs to reason within a physically grounded environment. Building on this representation, we formulate manipulation planning as a tree-structured search process~\cite{yao2023tree}, thereby extending both the temporal horizon and the spatial depth of VLM reasoning.

\subsection{Tree Construction for Manipulation Planning}
Classical manipulation planning frameworks rely on tree- or graph-structured search over symbolic task decompositions and continuous motion spaces, as exemplified by task-and-motion planning and behavior trees~\cite{doi:10.1126/scirobotics.abd9461}. More recent LLM-guided approaches, such as Tree-Planner~\cite{hu2024treeplannerefficientcloselooptask} and Prime the Search~\cite{Lee_2025}, improve search efficiency by refining candidate actions or warm-starting geometric planning using language priors. However, node feasibility in these methods is still assessed using semantic or geometric heuristics, leaving them vulnerable to physically infeasible branches that cannot be detected without explicit simulation.

STEP Planner~\cite{zhou2025stepplannerconstructingcrosshierarchical} further introduces a hierarchical subgoal tree to structure planning. Nevertheless, its feasibility evaluation relies on LLM-based semantic consistency, which remains susceptible to hallucinations when latent physical constraints (e.g., collisions, reachability limits, or support conditions) are not explicitly encoded in language.

In contrast, our method grounds both tree construction and search directly within a physics world model, evaluating node feasibility via simulated dynamic interaction rather than symbolic or purely geometric reasoning. Moreover, EToT employs \textit{Reflective Branching} to propose physically motivated revisions—such as obstacle relocation or action reordering—that may not be present in the original instruction but are essential for successful real-world execution.

\section{Methodology}

We propose the Embodied Tree of Thoughts (EToT) framework (Fig.~\ref{fig:etot}), which extends the reasoning capabilities of Vision-Language Models (VLMs) by coupling them with a physics-based embodied world model and a planning-tree search mechanism. The proposed methodology is organized around the following core questions:
(1) How are robot action skills formally defined? (Sec.~\ref{sec:action})
(2) How is a high-fidelity and interactive digital twin constructed? (Sec.~\ref{sec:Real2Sim})
(3) How is a world-model-grounded planning tree constructed and searched? (Sec.~\ref{sec:plan-tree})

\subsection{Action Skills Set}
\label{sec:action}

In this work, we focus on the task planning aspect of robotic manipulation. To standardize the planning process and facilitate the logical structuring of action sequences by the VLM, we model robot skills as a set of discrete semantic action primitives. Specifically, we define the following five action skills:
\[
\begin{cases}
\texttt{[PICK UP, obj]} \\
\texttt{[PUT ON, surface]} \\
\texttt{[PUT INTO, container]} \\
\texttt{[OPEN, obj]} \\
\texttt{[CLOSE, obj]}
\end{cases}
\]

Each action skill is implemented through a dedicated API. For the \textbf{PICK UP} action, we employ AnyGrasp~\cite{fang2023anygrasp} to estimate feasible grasp poses and the corresponding gripper width from RGB-D observations. The \textbf{PUT ON} action places the grasped object onto a specified supporting surface, while the \textbf{PUT INTO} action deposits the object into a designated container. The \textbf{OPEN} and \textbf{CLOSE} actions operate on articulated objects and are executed using manually scripted control primitives.

In our task setting, we do not explicitly model multiple candidate poses for most \textbf{PICK UP} or \textbf{PUT ON/INTO} actions. However, for objects that afford multiple distinct grasping orientations and for which identifying a feasible pose is nontrivial, we provide additional pose-specific configurations, denoted as \textbf{[PICK UP, obj] (POSE)}, to enable disambiguation among alternative grasp strategies.

\subsection{Embodied World Model Construction}
\label{sec:Real2Sim}
We construct an embodied world model using an efficient scene reconstruction pipeline that, in most cases, requires only a single RGB-D observation. Given an input RGB-D frame, we apply SAM-3~\cite{carion2025sam3segmentconcepts} to extract object masks from the RGB image, which are then processed by SAM-3D-Objects~\cite{sam3dteam2025sam3d3dfyimages} to generate textured object meshes. To recover metric scale, we employ the size estimation module of DexSim2Real$^{2}$~\cite{jiang2024dexsim2real2buildingexplicitworld} to produce scaled meshes from the RGB-D input. The scaled meshes, together with the RGB-D data and masks, are then passed to FoundationPose~\cite{wen2024foundationpose} for object pose estimation. Finally, the reconstructed meshes are imported into the OmniGibson~\cite{li2024behavior1k} simulator to generate an aligned and interactive digital twin of the physical scene.

For scenes containing articulated objects, an additional RGB-D frame captured from a different viewpoint with the object in an alternative kinematic state (e.g., open versus closed) is required. These multi-state observations are processed using DexSim2Real$^{2}$~\cite{jiang2024dexsim2real2buildingexplicitworld} to recover the articulated structure. Further implementation details are provided in the appendix.

\subsection{Planning Tree Construction and Searching.}
\label{sec:plan-tree}

As illustrated in Fig.~\ref{fig:etot}, we construct a world-model–grounded planning tree to support manipulation planning. The planning process consists of two key modules: Priori Branching and Reflective Branching. Priori Branching is responsible for generating the initial planning tree; however, the resulting branches may be invalid or unsafe due to incomplete physical reasoning. To address these issues, Reflective Branching analyzes simulated execution outcomes, identifies failure causes, and dynamically generates revised branches during the tree search process. The overall planning procedure is summarized in Alg.~\ref{alg:planning_tree_symbolic}.

\subsubsection{Priori Branching}

In the initial stage, the VLM analyzes the scene and task instruction to construct a preliminary planning tree grounded in object instances and candidate interaction modes.

\textit{Scene Parsing and Task Understanding.}
        The VLM takes as input an RGB image of the scene together with the task instruction. It extracts object-level information and infers relevant spatial relationships, including relational predicates (e.g., ``the pen is on the drawer'') and articulation states (e.g., ``the drawer is closed'').

\textit{Candidate Branches Generation.}  
        Based on the parsed scene representation, the VLM generates multiple candidate planning branches, each corresponding to a complete action sequence from the root to a leaf. The model is encouraged to explicitly branch at decision points where multiple feasible action choices exist. As illustrated in Fig.~\ref{fig:etot}, we consider two categories of branching:
        \begin{itemize}
            \item \textit{Instance-level branching:} Branches differ in object selection, for example, 
            (1) [PICK UP, pen], [PUT INTO, holder~1];  
            (2) [PICK UP, pen], [PUT INTO, holder~2].  
            \item \textit{Manipulation-parameter branching:} Branches differ in grasp configurations, for example,  
            (1) [PICK UP, holder] (Horizontally);  
            (2) [PICK UP, holder] (Vertically).  
        \end{itemize}
    
\textit{Initial Planning Tree Construction.}  
        Each action in a candidate branch is inserted as a node in the planning tree, and each complete branch forms a path from the root to a leaf. Branches sharing common action prefixes are merged to produce a compact, non-redundant tree representation.

\begin{algorithm}
\caption{Planning Tree Construction and Search}
\label{alg:planning_tree_symbolic}

\textbf{Symbols:}
 $\mathbf{T}$: planning tree,\;
 $\mathbf{Q}$: search queue,\;
 $\mathbf{N}$: tree node,\;
 $\mathbf{A}$: node action,\;
 $\mathcal{I}$: simulator-rollout image set,\;
 $\mathbf{R}$: action evaluation result,\;
 $\mathbf{P}$: extracted paths,\;
 $\mathbf{B}$: reflective candidate branch,\;
 $\mathcal{N}_{\text{merged}}$: nodes in $\mathbf{B}$ that overlap with existing tree,\;
 $\mathbf{n}_{\text{new}}$: first non-overlapping node produced during branch merging

\textbf{Input:} Scene image $I$, task instruction $t$, simulator $\mathcal{S}$  

\textbf{Output:} Feasible task plan $\pi^{*}$

\begin{algorithmic}[1]

\State $\mathbf{T} \gets \text{PrioriBranching}(I, t)$ 
\State $\mathcal{Q} \gets [\mathbf{T}.\text{root}.\text{children}]$

\While{$\mathbf{Q} \neq \emptyset$}

    \State $\mathbf{N} \gets \mathbf{Q}.\mathrm{pop}()$
    \State $\mathcal{I} \gets \mathcal{S}.\mathrm{execute}(\mathbf{N.A})$
    \State $\mathbf{R} \gets \mathrm{VLMJudge}(\mathbf{N.A}, \mathcal{I})$

    \If{$\mathbf{R} = \textit{Success}$}
        \If{$\mathbf{N}.\mathrm{children} = \emptyset$} \Comment{Leaf reached}
            \State \Return $\pi^{*} \gets \mathrm{extractPath}(\mathbf{N})$
        \Else
            \State $\mathbf{Q}.\mathrm{pushAll}(\mathbf{N}.\mathrm{children})$
        \EndIf

    \Else
        \Comment{Failure: refine via reflective branching}

        \State $\mathbf{P} \gets \mathrm{extractPaths}(\mathbf{N})$

        \State $\mathbf{T}.\mathrm{removeSubtree}(\mathbf{N})$

        \ForAll{$p \in \mathbf{P}$}
            \State $\mathbf{B} \gets \textbf{ReflectiveBranching}(\mathcal{I}, \mathbf{N}, p)$

            \State $(\mathcal{N}_{\text{merged}},\; \mathbf{n}_{\text{new}}) 
                \gets \mathbf{T}.\mathrm{mergeBranch}(\mathbf{B})$

            \If{$\mathbf{n}_{\text{new}} \neq \textit{None}$ \textbf{and} $\neg\mathrm{nInQ}(\mathcal{N}_{\text{merged}}, \mathbf{Q})$}
                \State $\mathbf{Q}.\mathrm{push}(\mathbf{n}_{\text{new}})$
            \EndIf
        \EndFor

    \EndIf
\EndWhile

\State \Return Task Planning Failed

\end{algorithmic}
\vspace{-0.2em}
\end{algorithm}

\subsubsection{Tree Searching and Reflective Branching}
We adopt a breadth-first search (BFS) strategy to traverse the planning tree layer by layer. When combined with reflective branching and dynamic plan revision, this strategy enables efficient discovery of feasible task plans while preserving both robustness and reasoning depth. During the search process, each node is evaluated sequentially along a branch: if the action associated with the current node is deemed feasible by the VLM, the search proceeds to the subsequent node; otherwise, the process transitions into the Reflective Branching stage. As illustrated in Fig.~\ref{fig:etot}, Reflective Branching consists of the following two procedures:

\begin{itemize}

\item \textit{Failure Detection.}
In real-world manipulation, successful execution requires not only achieving the intended task objective but also avoiding unintended disturbances or damage to surrounding objects. To assess both correctness and safety, the VLM compares the states of all objects before and after each simulated execution. An execution is classified as unsafe if it induces environmental changes that cannot be easily recovered using the available action skills, such as a tennis ball rolling off the table or a pen falling and sliding beyond the robot’s reachable workspace.

\item \textit{Tree Correction and Expansion.}  
Upon detecting execution failures or undesirable side effects, the VLM diagnoses the underlying cause and proposes a corrective strategy. In this work, we consider two primary categories of corrections.  
\textbf{(i) Collision-induced disturbances:} If the planned motion would result in collisions with nearby objects and alter their states, the corrective strategy first relocates the affected objects to safe positions before reattempting the original action.    
\textbf{(ii) Ordering-related conflicts:} Some failures arise from improper action ordering. For example, in Task~6 (Sec.~\ref{sec:task_design}), inserting the apple into the holder prior to relocating the holder prevents subsequent grasping of the holder. In such cases, the corrective strategy revises the branch by reordering the relevant actions, such as placing the holder at the target location before inserting the apple.
\end{itemize}

\subsection{Construction of closed-loop system}
\label{sec:closed-loop}

We further develop a closed-loop execution framework that incorporates real-robot feedback. After each action, the VLM evaluates the execution outcome using real camera observations in a manner consistent with simulation. Upon detecting a failure, the system reconstructs the current scene as a new initial state and uses it as the root for replanning. The updated state and original task instruction are then used to regenerate a new planning tree, followed by another round of tree construction and search. This design enables continuous feedback-driven correction during real-world execution.

\section{Experiments}

In this section, we present the experimental setup (Sec.~\ref{sec:exp_setup}) and task design (Sec.~\ref{sec:task_design}), followed by a detailed analysis of the results (Sec.~\ref{experiment results}) and an ablation study examining the contribution of each component of our framework(Sec.~\ref{ablation}). We further investigate the feasibility of accelerating the inference process in Sec.~\ref{para}.

\subsection{Experimental Setup}
\label{sec:exp_setup}

\begin{figure}[t]
    \centering
    \includegraphics[width=0.5\textwidth]{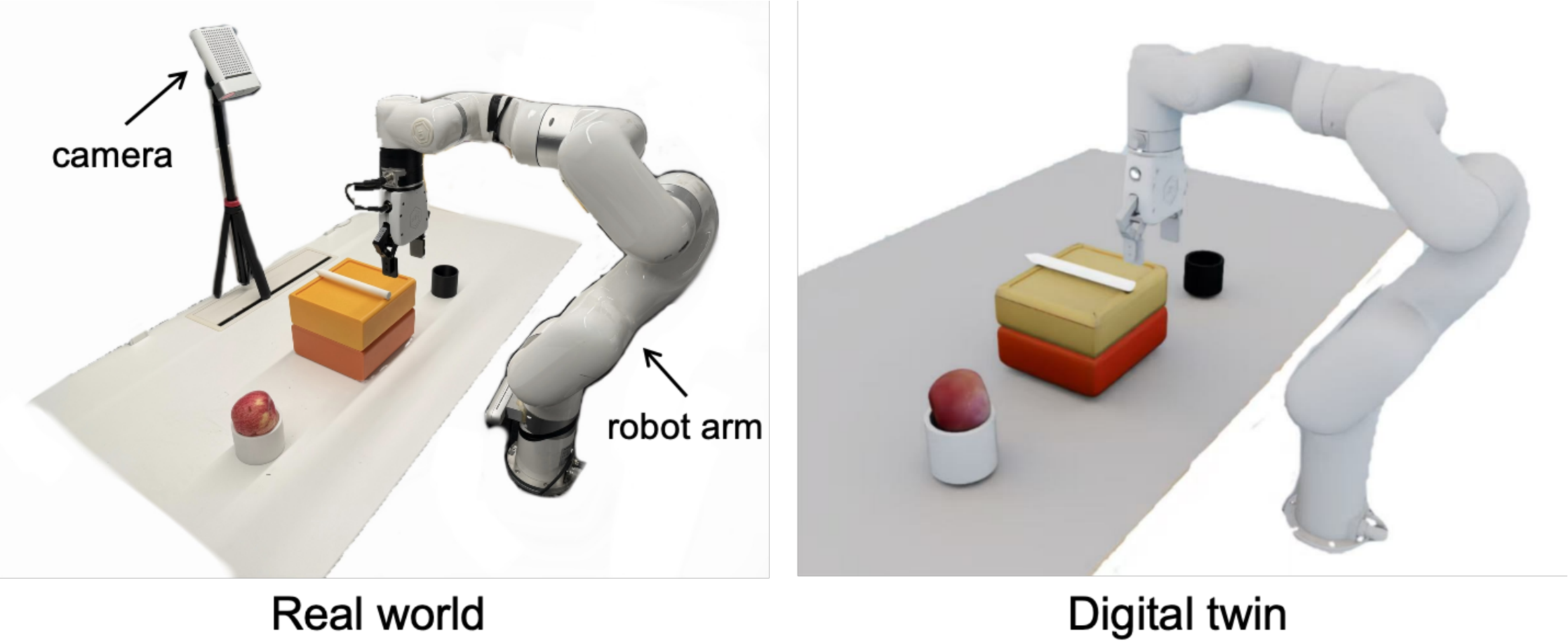}
    \caption{\textbf{Experimental scenarios in real world and simulation.}}
    \label{fig:robot}
\end{figure}

\textbf{Hardware.} As illustrated in Fig.~\ref{fig:robot}, all experiments are conducted using an xArm6 6DoF robot manipulator with a parallel-jaw gripper (1 DoF) and an Azure Kinect DK RGB-D camera.

\textbf{Baselines.} We compare our method against three baselines.
\textbf{ReKep} \cite{huang2024rekep} extracts visual keypoints and integrates them with VLM-guided reasoning to generate constraint-based task specifications for manipulation planning.
\textbf{ReKep w/ CoT} \cite{wei2022chain} augments ReKep with explicit chain-of-thought prompting to encourage more deliberate and fine-grained reasoning. We design concise multi-step prompts that guide the VLM to analyze task dependencies and anticipate potential failure modes, resulting in deeper, more logically grounded action plans.
We augment ReKep w/ CoT with an oracle variant of the reflective mechanism proposed in Reflect~\cite{liu2023reflect}, yielding \textbf{Reflect$^{*}$}. This baseline employs the same VLM-based execution evaluation module to assess action outcomes. Upon detecting a failure, an oracle-style reflection step is invoked, in which a valid recovery plan is manually specified using the available action primitives whenever such a plan exists.

In our implementation, we standardize all configurations across methods except for the task planning component, including keypoint extraction and the action primitive set. All methods are provided with manually annotated, high-precision keypoints to ensure a fair comparison. Due to the heterogeneity in the definitions of action primitives across prior works (e.g., the original ReKep prompt does not include operations on articulated objects), we employ a manual post-processing procedure to map the plans generated by each method into our unified API. Specifically, semantic plans produced by ReKep that involve actions such as closing a drawer via keypoint specification are manually converted into our corresponding \texttt{CLOSE} API. Please refer to the appendix~\ref{case analysis} for the detailed procedures of these baselines.

\textbf{Metrics.} We evaluate each method by \textit{task success rate}, defined as the proportion of trials that achieve the specified goal without causing harmful changes to the environment.
    
\textbf{Implementation Details.} All baselines and our proposed method are implemented using GPT-4o~\cite{achiam2023gpt} as the underlying vision–language model. The camera is mounted at a fixed oblique viewing angle. For each task, all experiments are repeated for 10 independent trials.

\subsection{Task Design}
\label{sec:task_design}

We design a suite of seven manipulation tasks to systematically evaluate four fundamental capabilities of robotic planning:
\textit{(a)} awareness of object manipulability,
\textit{(b)} understanding of three-dimensional spatial relationships,
\textit{(c)} prediction of physical dynamics, and
\textit{(d)} robustness to external disturbances with automatic recovery.

Tasks involving three or fewer action steps are categorized as short-horizon tasks (Tasks~1–4), whereas tasks requiring more than three actions are classified as long-horizon tasks (Tasks~5–7). In addition, we introduce a disturbance-aware task, where disturbances correspond to human-induced interference during task execution. An overview of all tasks is illustrated in Figure~\ref{fig:tasks}, and detailed descriptions of the task design are provided in Appendix~\ref{task_design}.

\textbf{Task 1: Open the door of the microwave oven.} (\textit{b, c}) Directly opening the door would cause the tennis ball on the desk to fall; the ball must therefore be relocated prior to door actuation.

\textbf{Task 2: Reorient a pen and place it into a holder.} (\textit{c}) The pen is unstable when inserted into the black holder and topples due to insufficient support, requiring placement into the white holder for stable insertion.

\textbf{Task 3: Pick up the holder horizontally or vertically} (\textit{a}) (\textit{a}) A horizontal side grasp induces slippage due to low surface friction and near-maximum gripper width, necessitating a top-down grasp strategy.

\begin{figure*}[!t]
    \centering
    \includegraphics[width=0.9\textwidth]{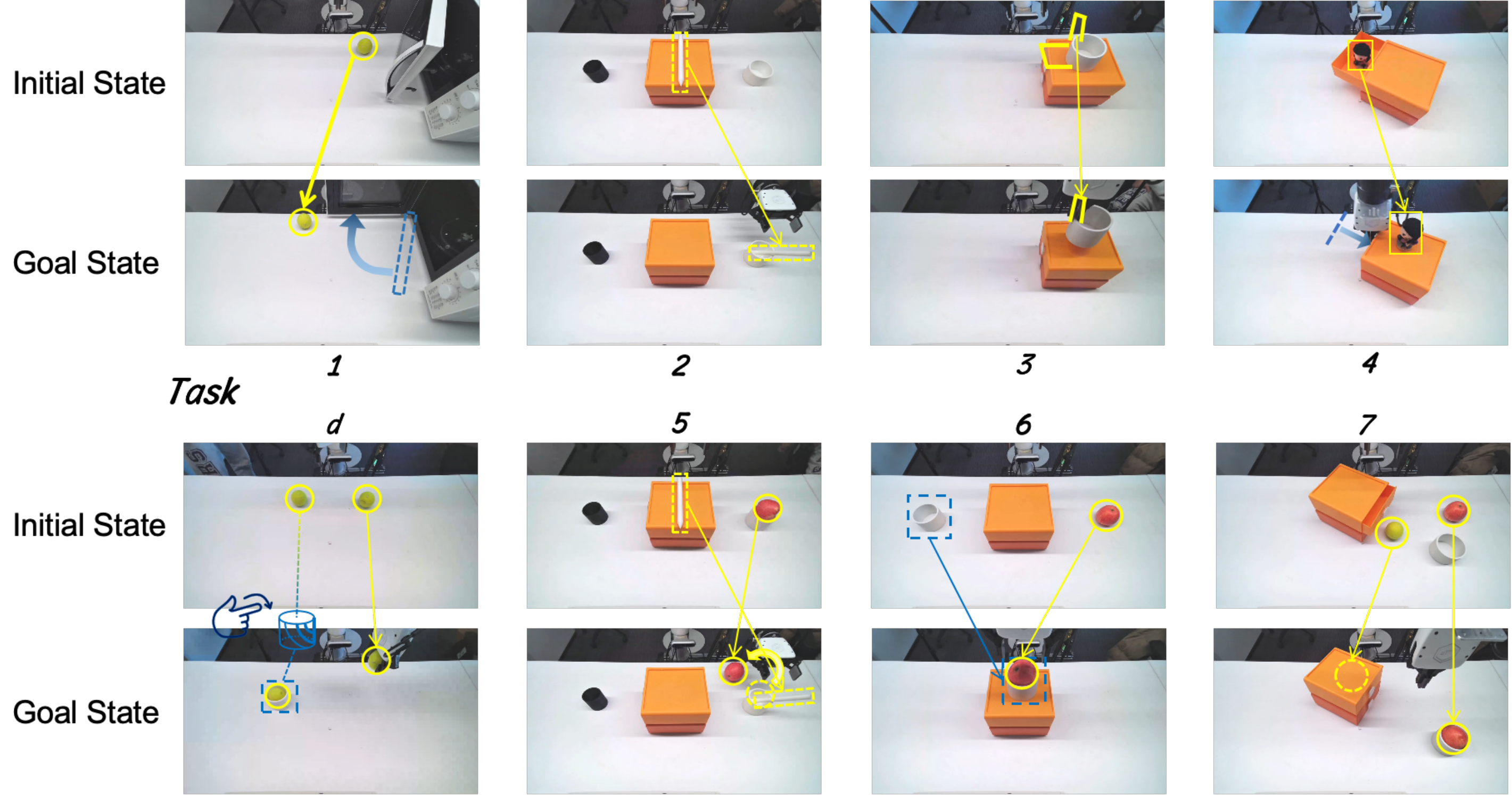}
    \caption{\textbf{Schematic diagrams of all tasks, including the initial states and the correct goal states, with boxes and arrows indicating the position changes of key objects. We provide detailed plan trees in the appendix(Fig.~\ref{fig:tasks1-4}, \ref{fig:tasksd5-7}).}}
    \label{fig:tasks}
    \vspace{-3mm}
\end{figure*}

\textbf{Task 4: Close the drawer.} (\textit{b, c}) The toy inside the drawer exceeds the clearance for closure, requiring relocation to a safe position before closing the drawer.

\textbf{Disturbance Task: Pick up a tennis ball.} (\textit{a, d}) After human-induced disturbance renders the original target ungraspable, the robot must detect failure and replan to grasp the alternative ball.

\textbf{Task 5: Reorient a pen and place it into a holder.} (\textit{b, c}) An apple on the white holder obstructs insertion and causes rebound, requiring removal of the apple prior to placing the pen.

\textbf{Task 6: Place the apple and the holder on the drawer, with the apple inside the holder.} (\textit{a, b}) Inserting the apple before placing the holder blocks the required top-down grasp; the holder must be positioned first, followed by insertion of the apple.

\textbf{Task 7: Put the apple and the tennis ball in either the drawer or the pen holder, together or separately. Ensure the drawer is closed.} (\textit{a, b, c}) The apple violates the drawer height constraint and the tennis ball initially occludes access, requiring relocation of the ball prior to placing the apple in the holder and the ball in the drawer.

\subsection{Experimental Results and Discussion}
\label{experiment results}
Table~\ref{table_1} summarizes the success rates of our method in comparison with all baselines across the seven tasks. Our approach consistently outperforms the baselines on every task and achieves the highest overall average success rate of \textbf{88.8\%}.

ReKep generates constraint-based plans grounded in VLM-predicted keypoints; however, its limited fine-grained three-dimensional spatial reasoning and physical dynamics prediction lead to poor performance across most tasks, resulting in the lowest overall success rate. ReKep w/ CoT augments this process with additional chain-of-thought reasoning, yielding noticeable improvements on Tasks~1 and~4. For instance, in Task~1, it correctly identifies the safety risk that opening the microwave door would displace the tennis ball and generates a plan that first relocates the ball. Nevertheless, its performance gains on the remaining tasks remain limited.

Reflect$^{*}$ further incorporates reflective planning and demonstrates clear improvements on Task~3, where an initial failed horizontal grasp can be corrected by a subsequent top-down attempt. It also performs comparably to our method on the Disturbance Task by adapting its plan after execution failures. However, even with oracle-level reflective reasoning, Reflect provides no benefit in tasks involving irreversible failures (e.g., Task~6, where the apple cannot be recovered once it falls into the holder). In contrast, our method leverages a physics-based world model to simulate candidate actions in advance, identify potential risks prior to execution, and proactively generate safer plans to avoid such failures.

On long-horizon tasks, all three baselines exhibit limited performance, with none exceeding a 50\% success rate, due to the increased complexity of spatial relations, physical constraints, and extended branching factors. By contrast, our method exploits the predictive capability of the physics world model and performs a more comprehensive VLM-guided tree search, maintaining high success rates on Tasks~5–7 (8/10, 9/10, and 7/10, respectively).

\subsection{Ablation}
\label{ablation}
\begin{table*}[!t]
\centering
\renewcommand{\arraystretch}{1.4}
\setlength{\tabcolsep}{12pt}
\begin{tabular}{c|cccc|c|ccc|c}
\thickhline
\textbf{Success Rate} & \multicolumn{5}{c|}{\textbf{Short Tasks}} & \multicolumn{3}{c|}{\textbf{Long Tasks}} & \multirow{2}{*}{\textbf{Avg}}\\
\cline{2-6} \cline{7-9}
 & Task1 & Task2 & Task3 & Task4  & Disturbance & Task5 & Task6 & Task7 & \\
\hline
ReKep\cite{huang2024rekep}   & 0/10 & 5/10 & 4/10 & 0/10 & 0/10 & 0/10 & 4/10 & 0/10 & 16.3\%\\
ReKep w/ CoT\cite{wei2022chain} & 8/10 & 7/10 & 0/10 & 8/10 & 0/10 & 1/10 & 5/10 & 2/10 & 38.8\%\\
Reflect$^{*}$~\cite{liu2023reflect}  & 8/10 & 7/10 & 8/10 & \underline{10/10} & \underline{10/10} & 2/10 & 5/10 & 3/10 & 66.3\%\\
\textbf{EToT}   & \textbf{9/10} & \textbf{9/10} & \textbf{9/10} & \textbf{\underline{10/10}} & \textbf{\underline{10/10}} & \textbf{8/10} & \textbf{9/10} & \textbf{7/10} & \textbf{88.8\%}\\
\thickhline
\end{tabular}
\caption{\textbf{Comparison of Success Rates Across Short- and Long-Horizon Tasks.}
We compare different planning baselines on seven manipulation tasks spanning short- and long-horizon scenarios, including an additional disturbance-aware task. The highest success rate for each task is highlighted in bold, and entries achieving the same highest value are additionally underlined.}
\label{table_1}
\end{table*}

\begin{table*}[!t]
\centering
\renewcommand{\arraystretch}{1.4}
\setlength{\tabcolsep}{12pt}
\begin{tabular}{c|cccc|c|ccc|c}
\thickhline
\textbf{Success Rate} & \multicolumn{5}{c|}{\textbf{Short Tasks}} & \multicolumn{3}{c|}{\textbf{Long Tasks}} & \multirow{2}{*}{\textbf{Avg}}\\
\cline{2-6} \cline{7-9}
 & Task1 & Task2 & Task3 & Task4  & Disturbance & Task5 & Task6 & Task7 & \\
\hline
w/o Priori   & \underline{9/10} & 5/10 & 4/10 & \underline{10/10} & 5/10 & 5/10 & \underline{9/10} & 2/10 & 61.3\%\\
w/o Reflective  & 0/10 & \underline{9/10} & \underline{9/10} & 0/10 & \underline{10/10} & 0/10 & 0/10 & 0/10 & 35.0\%\\
w/o Replan  & \underline{9/10} & \underline{9/10} & 8/10 & \underline{10/10} & 0/10 & \underline{8/10} & \underline{9/10} & \underline{7/10} & 75.0\%\\
\hline
w/ VGM & 3/10 & 5/10 & 4/10 & 0/10 & 5/10 & 0/10 & 0/10 & 0/10 & 21.3\%\\
\thickhline
\textbf{Full}  & \textbf{\underline{9/10}} & \textbf{\underline{9/10}} & \textbf{\underline{9/10}} & \textbf{\underline{10/10}} & \textbf{\underline{10/10}} & \textbf{\underline{8/10}} & \textbf{\underline{9/10}} & \textbf{\underline{7/10}} & \textbf{88.8\%}\\
\thickhline
\end{tabular}
\caption{\textbf{Ablation Study.} We investigate the contributions of Priori Branching, Reflective Branching, and the Replanning mechanism by selectively removing each component. In addition, we assess a variant that replaces the physics simulator with a video generation model (VGM) as the world model.}
\label{table_2}
\end{table*}

Table~\ref{table_2} summarizes the ablation results assessing the contributions of Priori Branching, Reflective Branching, and real-world Replanning, as well as the impact of replacing the physics simulator with a VGM~\cite{jimeng}. For VGM-based planning, the real camera image at the start of each action is used as the initial frame, and the VGM is conditioned on the corresponding action instruction to generate a predicted execution video. To avoid temporal error accumulation, each action is reinitialized from a real image rather than using generated frames.

Removing either Priori Branching or Reflective Branching results in substantial performance degradation.Without Priori Branching, the VLM produces a single-branch search tree and frequently commits to incorrect high-level decisions in multi-path tasks (e.g., Tasks~2, 5, and 7), preventing recovery.  Without Reflective Branching, the system loses its ability to diagnose and revise failures in simulation, leading to sharp drops on Tasks~1, 4, and all long-horizon tasks, with the overall success rate reduced to 35.0\%. Disabling real-world replanning (w/o Replan) preserves performance on short-horizon tasks but completely fails under disturbances (0/10), highlighting the importance of feedback-driven correction.

\begin{figure}
    \centering
    \includegraphics[width=0.9\linewidth]{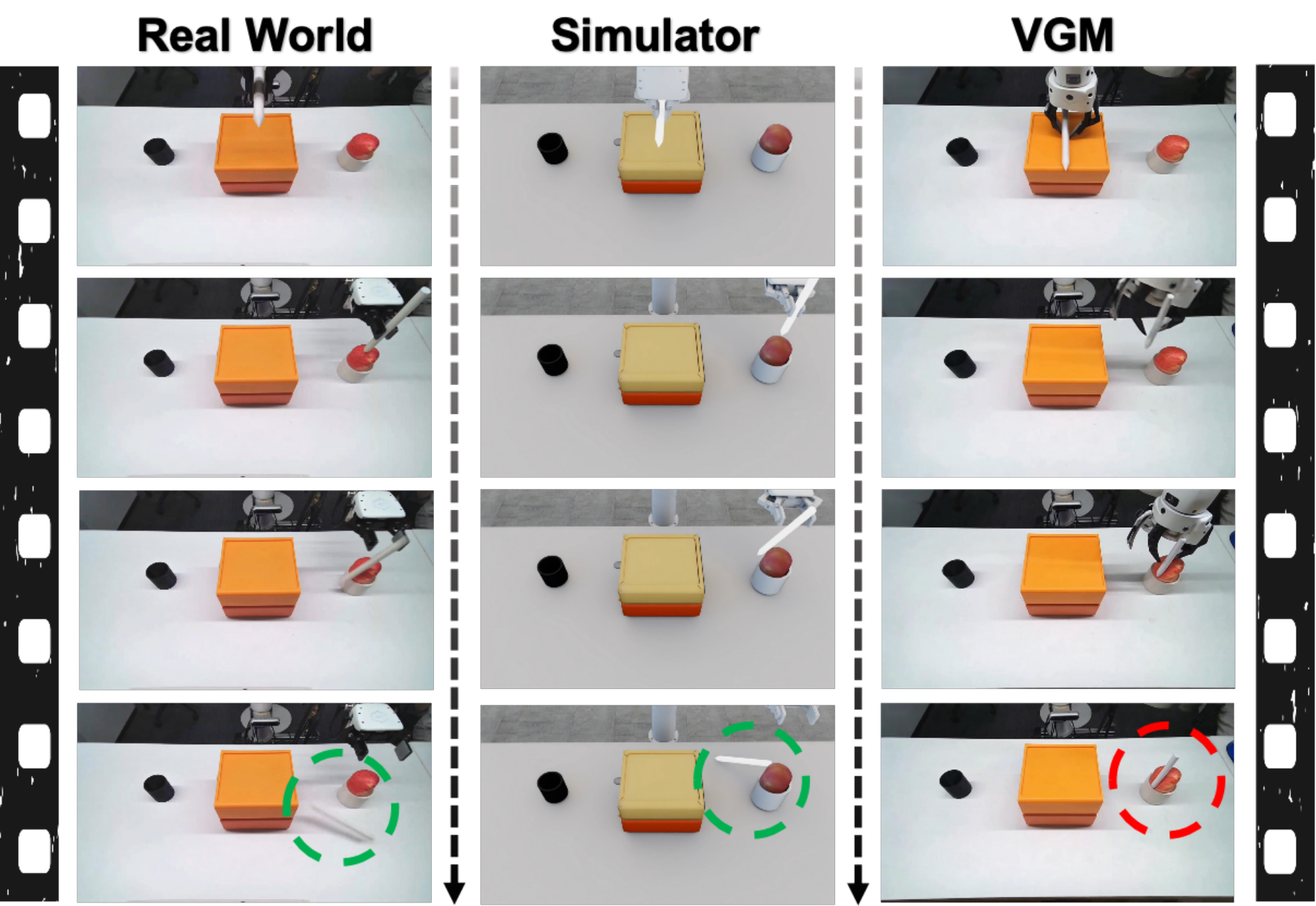}
    \caption{Comparison of scene evolution in the real world, the physics-based simulator, and the video generation model (VGM) for the action ``Put the pen into holder~2'' in Task~5 }
    \label{fig:wm}
    \vspace{-3mm}
\end{figure}

Replacing the physics simulator with a VGM further reduces the average success rate to 21.3\%, as the VGM lacks physically consistent prediction and reliable feasibility evaluation. As shown in Fig.~\ref{fig:wm}, in the ``put the pen into holder~2'' task, the simulator correctly predicts slippage after contact with the apple, whereas the VGM incorrectly depicts successful insertion.

Overall, the ablation study confirms that Priori Branching, Reflective Branching, physics-based simulation, and real-world replanning are all indispensable and complementary components of EToT.

\subsection{Failure case analysis}

\begin{figure}
    \centering
    \includegraphics[width=0.9\linewidth]{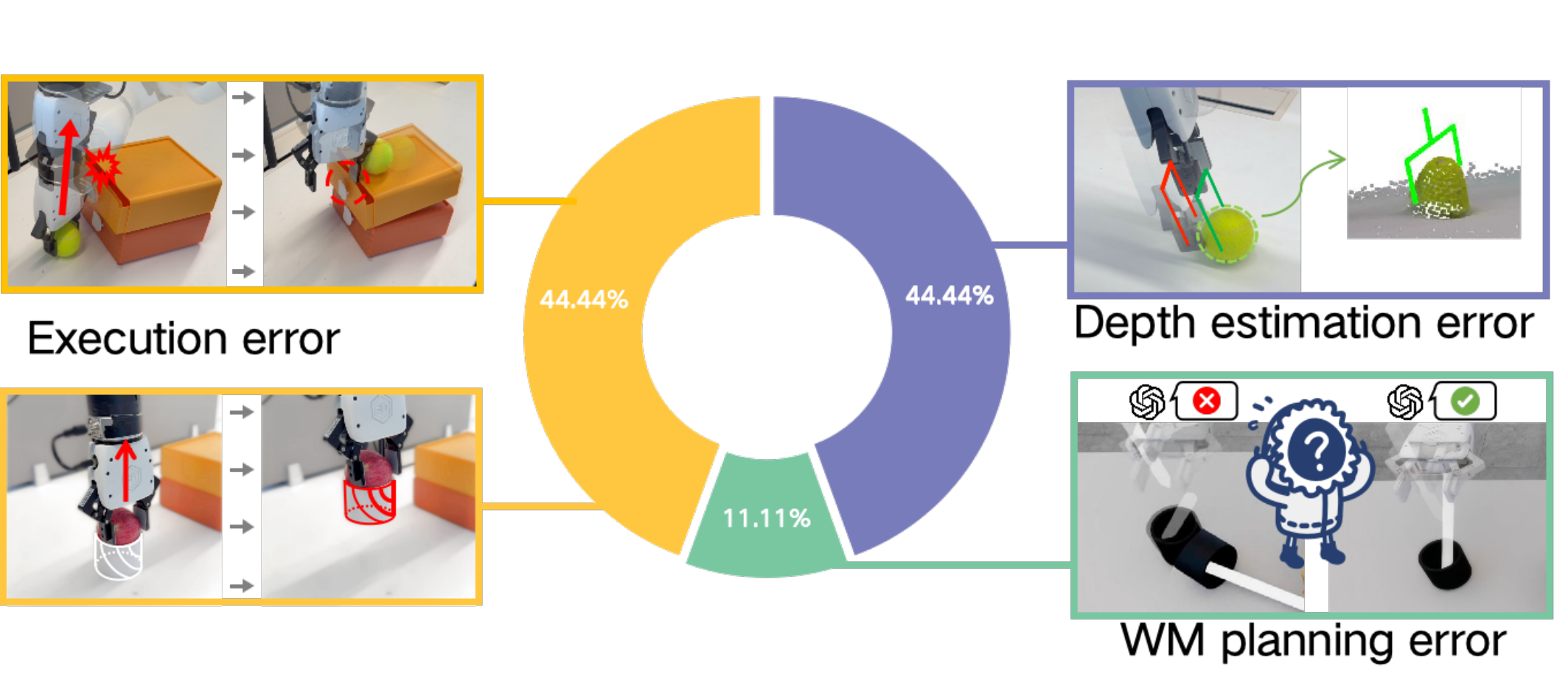}
    \caption{Failure analysis. Representative examples and the percentage of each failure type are shown.}
    \label{fig:failure}
    \vspace{-5mm}
\end{figure}

Figure~\ref{fig:failure} summarizes the primary failure modes observed in our real-world experiments, which can be categorized into execution errors, depth estimation errors, and world-model (WM) planning errors.
\textbf{Execution errors} (44.44\%) occur during physical interaction or motion execution, primarily due to unmodeled contact dynamics and control inaccuracies. In one case, the end-effector collides with the drawer while approaching the tennis ball, unintentionally closing the drawer and disrupting the subsequent opening action. In another case, an unintended downward force during gripper closure causes the apple to wedge against the holder, resulting in the apple and holder being lifted together and preventing further task execution.
\textbf{Depth estimation errors} (44.44\%) oarise from distorted or noisy depth measurements. As shown, although the nominal grasp pose (green) is geometrically valid, corrupted depth values shift the estimated pose to an incorrect location (red), leading to a failed grasp of the tennis ball.
\textbf{WM planning errors} (11.11\%) are caused by physically implausible predictions from the world model. In the illustrated example, inserting the pen should induce a tipping motion of the holder; however, the simulated trajectory incorrectly stabilizes the pen against the gripper, causing the planner to accept an infeasible action sequence.

\subsection{Parallel acceleration}
\label{Parallel acceleration}
\label{para}
The computational cost of exhaustive tree-based planning grows exponentially with the search depth and branching factor, leading to significant scalability challenges. To address this issue, we adopt a \textbf{multi-world parallelism} strategy, in which all newly generated child nodes pending evaluation are aggregated into a batch. Each child node initializes from the terminal state of its parent, and the corresponding actions are executed in parallel. This parallelization scheme reduces the inference time for Task~7 by approximately 50\%, demonstrating its effectiveness in improving computational efficiency.

\section{Conclusion}

This work presents Embodied Tree of Thoughts (EToT), a deliberative manipulation planning framework that integrates tree-structured search with a physics-based embodied world model. By jointly leveraging \textit{Priori Branching} and \textit{Reflective Branching}, EToT enhances physical reasoning, anticipates potential execution failures prior to real-world deployment, and generates robust and safe manipulation plans. Extensive experimental results demonstrate that EToT consistently outperforms existing approaches, with particularly significant advantages on complex long-horizon tasks where sequential dependencies and physical constraints are critical.

At present, the proposed framework is validated on tabletop manipulation scenarios using a fixed-base manipulator and a discrete set of pick-and-place primitives. Future work will focus on extending EToT to mobile manipulation settings and incorporating a richer repertoire of skills, such as pushing and pressing, to further improve the practicality, generality, and scalability of the framework for real-world robotic applications.

\bibliographystyle{IEEEtran}
\bibliography{main}

\clearpage
\appendix
\section{Appendix}

\subsection{Task design}
\label{task_design}
\begin{figure*}
    \centering
    \includegraphics[width=0.95\textwidth]{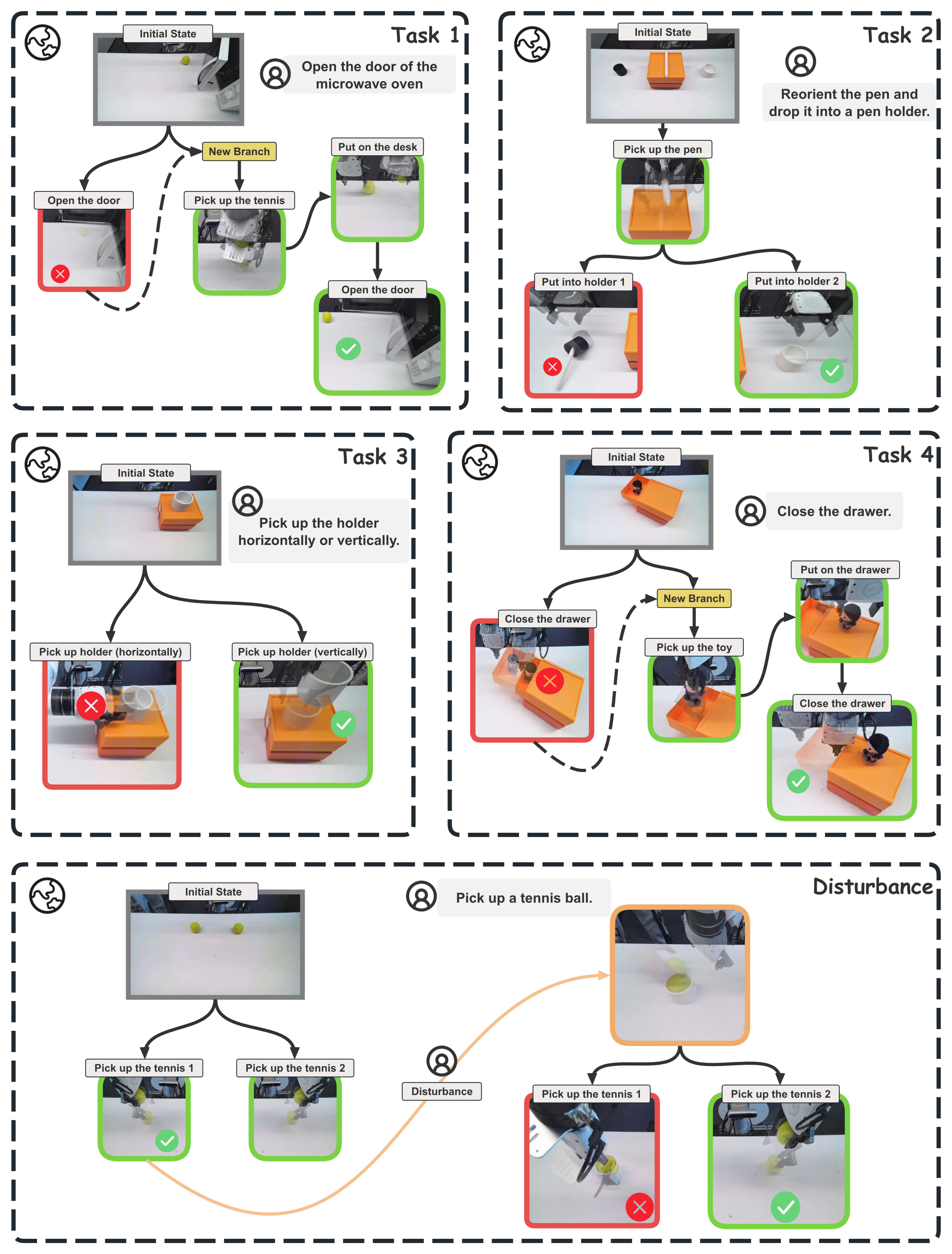}
    \caption{\textbf{Short-horizon tasks(Task1-4, Disturbance)}}
    \label{fig:tasks1-4}
\end{figure*}

\begin{figure*}
    \centering
    \includegraphics[width=1.0\textwidth]{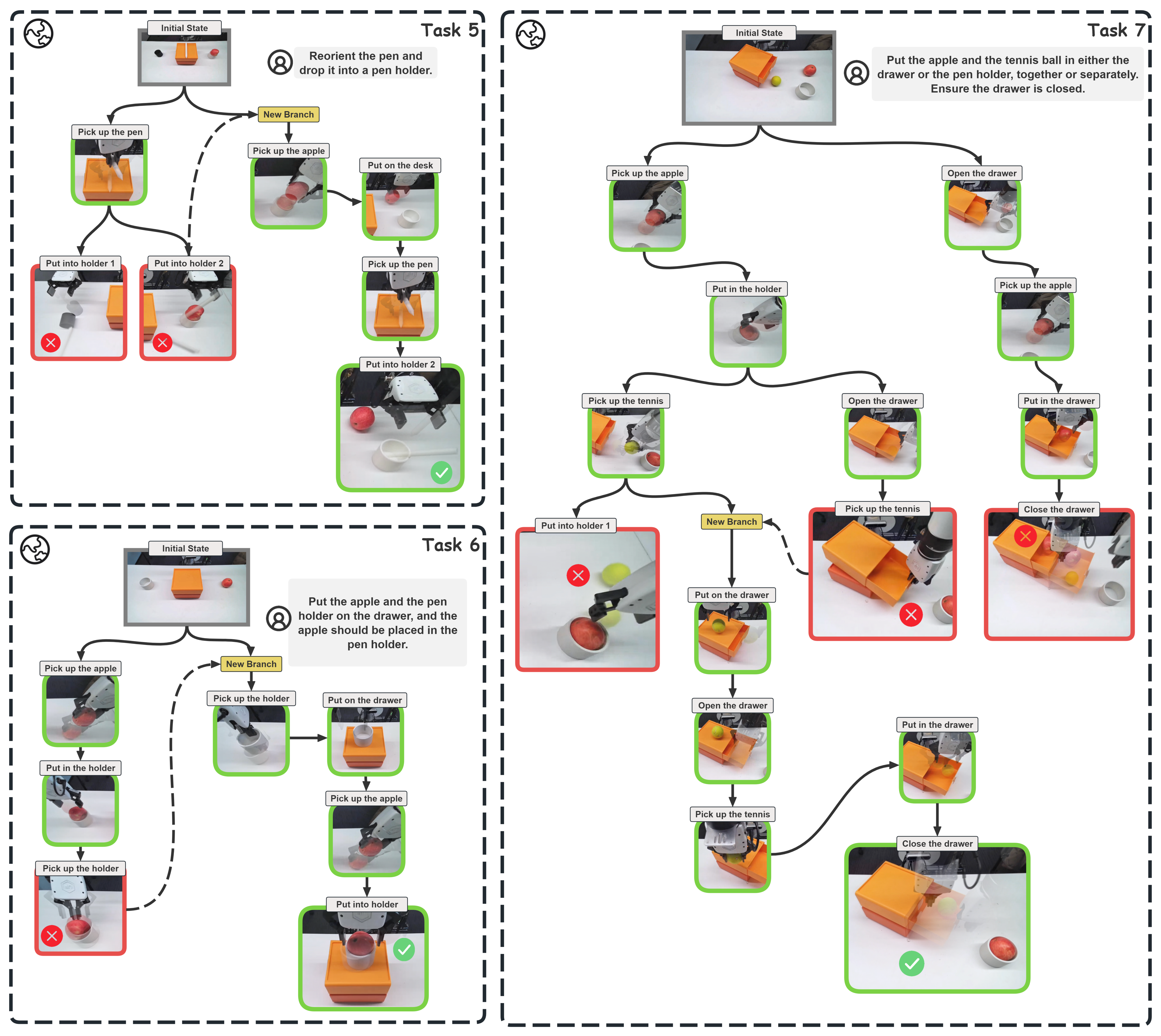}
    \caption{\textbf{Long-horizon tasks(Task5-7)}}
    \label{fig:tasksd5-7}
\end{figure*}

In this section, we elaborate on the scenario of each task, encompassing the inherent challenges and the correct solutions. Additionally, we provide schematic diagrams (Fig.~\ref{fig:tasks1-4}, \ref{fig:tasksd5-7}) illustrating the reasoning pathways for each task.

We evaluate four essential capabilities of robotic planning:  
\begin{itemize}
    \item \textit{(a)} object manipulability awareness,
    \item \textit{(b)} understanding of 3D spatial relations,
    \item \textit{(c)} prediction of physical dynamics,
    \item \textit{(d)} robustness to disturbances with automatic recovery.
\end{itemize}

\textbf{Task 1: Open the door of the microwave oven.} (\textit{b, c})  
A closed microwave and a tennis ball are placed on the desk surface. Directly opening the door would push the ball off the desk. The correct sequence of actions is to relocate the ball before opening the door.

\textbf{Task 2: Reorient a pen and place it into a holder.} (\textit{c})  
Two pen holders (black and white) are positioned on the left and right sides of the desk, with a pen lying on a drawer. Due to differences in shape and mass, placing the pen into the black holder causes it to fall, whereas the white holder ensures a stable placement. The correct procedure is to place the pen into the white holder.

\textbf{Task 3: Pick up the holder horizontally or vertically.} (\textit{a})  
A pen holder is placed on a drawer. Two grasp strategies are provided via annotated grasping keypoints. Because the holder is slightly wider than the gripper and has low friction, performing a horizontal side grasp results in slippage, whereas a top–down grasp succeeds.

\textbf{Task 4: Close the drawer.} (\textit{b, c})  
A drawer is open with a toy inside. The toy is slightly taller than the drawer compartment, preventing the drawer from closing if pushed directly. The correct plan is to move the toy to a safe location above the drawer before closing it.

\textbf{Disturbance Task: Pick up a tennis ball.} (\textit{a, d})  
Two tennis balls are initially graspable. After the robot commits to grasping one ball, a human-induced disturbance places that ball into a holder, making it ungraspable under the available action skill set. The correct behavior is to replan and grasp the other ball.

\textbf{Task 5: Reorient a pen and place it into a holder.} (\textit{b, c})  
This long-horizon task extends Task~2 by placing an apple on top of the white holder. Inserting the pen directly causes it to rebound off the apple. The correct plan is to first move the apple to a safe location, then place the pen into the white holder.

\textbf{Task 6: Place the apple and the holder on the drawer, with the apple inside the holder.} (\textit{a, b})  
A white holder (as used in Task~3), an apple, and a drawer are presented. Since the holder only supports a top–down grasp, placing the apple inside beforehand blocks the grasp approach and prevents subsequent manipulation. The correct sequence is to place the holder on the drawer first, then place the apple into the holder.

\textbf{Task 7: Put the apple and the tennis ball in either the drawer or the pen holder, together or separately. Ensure the drawer is closed.} (\textit{a, b, c})  
A slightly ajar drawer, a tennis ball, an apple, and a holder are available. The holder can accommodate either object but not both. The drawer has sufficient width for both objects, but the apple is too tall for the drawer to close. Thus, the correct allocation places the apple into the holder and the ball into the drawer. Additionally, because the ball is initially located directly below the drawer front, opening the drawer first blocks access to the ball. The correct sequence is to move the ball to a safe location on top of the drawer, open the drawer, and then place the ball inside.

\subsection{3D reconstruction}
\label{ap_3d}
We construct our 3D reconstruction pipeline with the objective of generating digital twins that are (i) aligned with real-world geometry, (ii) compatible with the OmniGibson simulator format, and (iii) produced using as few viewpoints, procedural steps, and human interventions as possible. Below, we describe the reconstruction method in detail, including input requirements for each stage and the points at which minimal manual assistance is needed.

\begin{enumerate}
    \item \textbf{Mask Acquisition.}
    SAM3~\cite{carion2025sam3segmentconcepts} can automatically segment target objects based on task-level instructions. Masks can be obtained either through the web-based demo—used in this work for convenience—or via local deployment. For scenes containing complex object stacks, segmentation errors may occur; in such cases, the user can specify keypoints to guide SAM3 toward more accurate mask extraction.
 
    \item \textbf{Initial 3D Model Generation.}
    We employ SAM-3D-OBJECTS~\cite{sam3dteam2025sam3d3dfyimages} to generate initial textured 3D models. A single-view RGB image (e.g., from our camera viewpoint) together with the corresponding object masks is used as input to produce mesh models in GLB format. However, the resulting meshes typically do not preserve real-world scale, which motivates the subsequent size-optimization step.

    \item \textbf{Size adjustment.}
    We adopt the size-optimization module of Dex. Given the mesh generated in the previous step, along with the RGB-D image, masks, and camera extrinsics, Dex automatically adjusts the scale of each model. Due to mesh deformation during generation, inaccuracies in depth measurements, or camera distortions, the optimized mesh may still deviate slightly from real dimensions. In such cases, minor manual corrections are required. This step yields mesh models accurately aligned with real-world object sizes.

    \item \textbf{Pose Estimation.}
    We compute each object’s pose in the camera coordinate frame using FoundationPose. For each object, we input the refined mesh model, RGB-D image, and mask. Directly using the GLB model with FoundationPose results in errors; we provide the necessary code modifications to resolve this issue. As with previous stages, stacked objects, camera distortions, and depth-map inaccuracies may adversely affect pose estimation, requiring limited manual adjustment when necessary.

    \item \textbf{Reconstruction of Articulated Objects.}
    For non-articulated rigid objects, the above pipeline using a single-view RGB-D image suffices to produce high-quality digital twins. Articulated objects, however, pose additional challenges: SAM-3D-Objects tends to generate a single unified mesh, necessitating substantial manual labor to segment components, clear internal cavities, and define joint locations and limits. We instead use Dex, which reconstructs articulated objects from two RGB-D views (frontal and lateral). This approach produces realistic geometry and textures, correct joint placement, complete internal structure, and artifacts suitable for subsequent robotic manipulation tasks.

    \item \textbf{Importing into the OmniGibson Simulator.}
    By default, OmniGibson generates collision volumes via convex hulls for all meshes. For non-convex objects, this may cause significant physical inaccuracies (e.g., a pen holder’s opening being sealed by the convex hull, preventing insertion of pens). For such geometrically non-convex meshes, we preprocess them with COACD to obtain a convex decomposition before importing them into the simulator.
\end{enumerate}

\subsection{Case Analysis}
\label{case analysis}

\begin{figure*}
    \centering
    \includegraphics[width=0.9\textwidth]{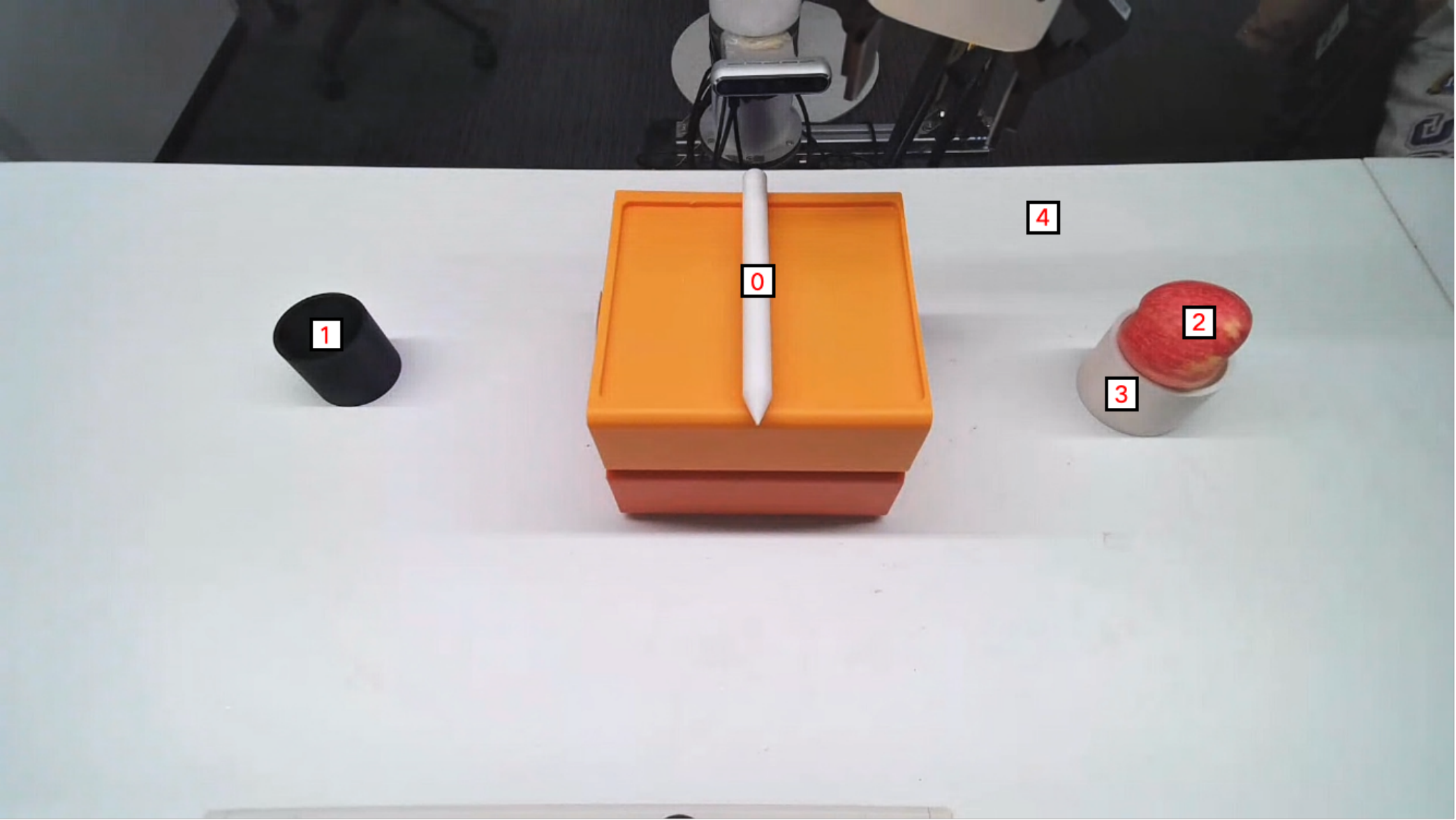}
    \caption{\textbf{Camera image with manually annotated key points.}}
    \label{task5points}
\end{figure*}

To help readers intuitively understand how the methods introduced in this paper are applied in task planning, we use Task~5 as an example to illustrate the planning process of each method and annotate the camera images at key steps, as shown in Figure~\ref{task5points}.

\subsubsection{ReKep}

\begin{figure*}
    \centering
    \includegraphics[width=0.8\textwidth]{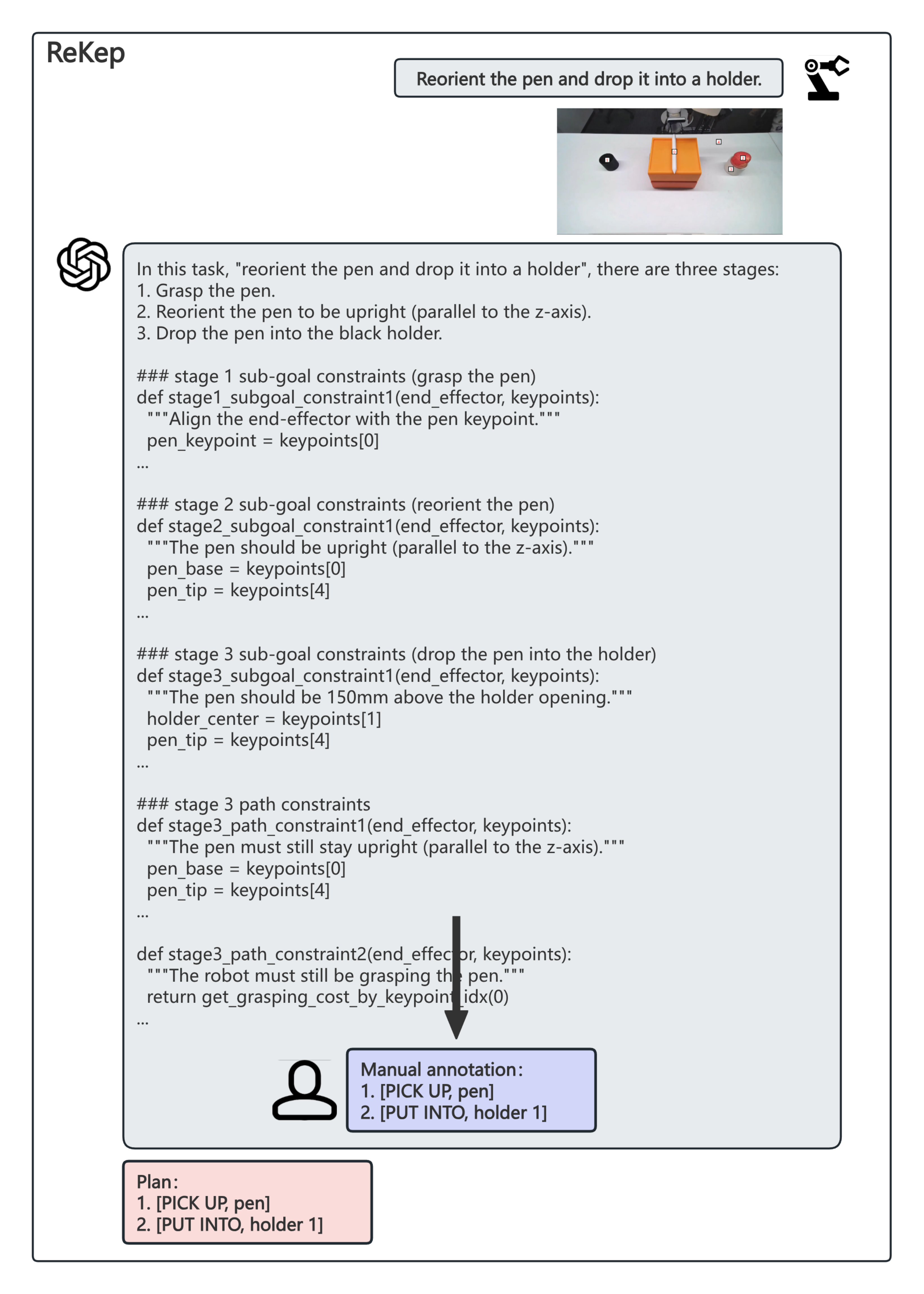}
    \caption{\textbf{Example of ReKep planning.}}
    \label{rekepcase}
\end{figure*}

Figure~\ref{rekepcase} shows an example of ReKep planning based on the annotated keypoint images and task instructions. ReKep first generates keypoint-based constraints for planning, after which we manually transform them into executable solutions. Its keypoint tracking and backtracking functions are implemented through manual supervision and intervention.

\subsubsection{ReKep w/ CoT}

\begin{figure*}
    \centering
    \includegraphics[width=0.95\textwidth]{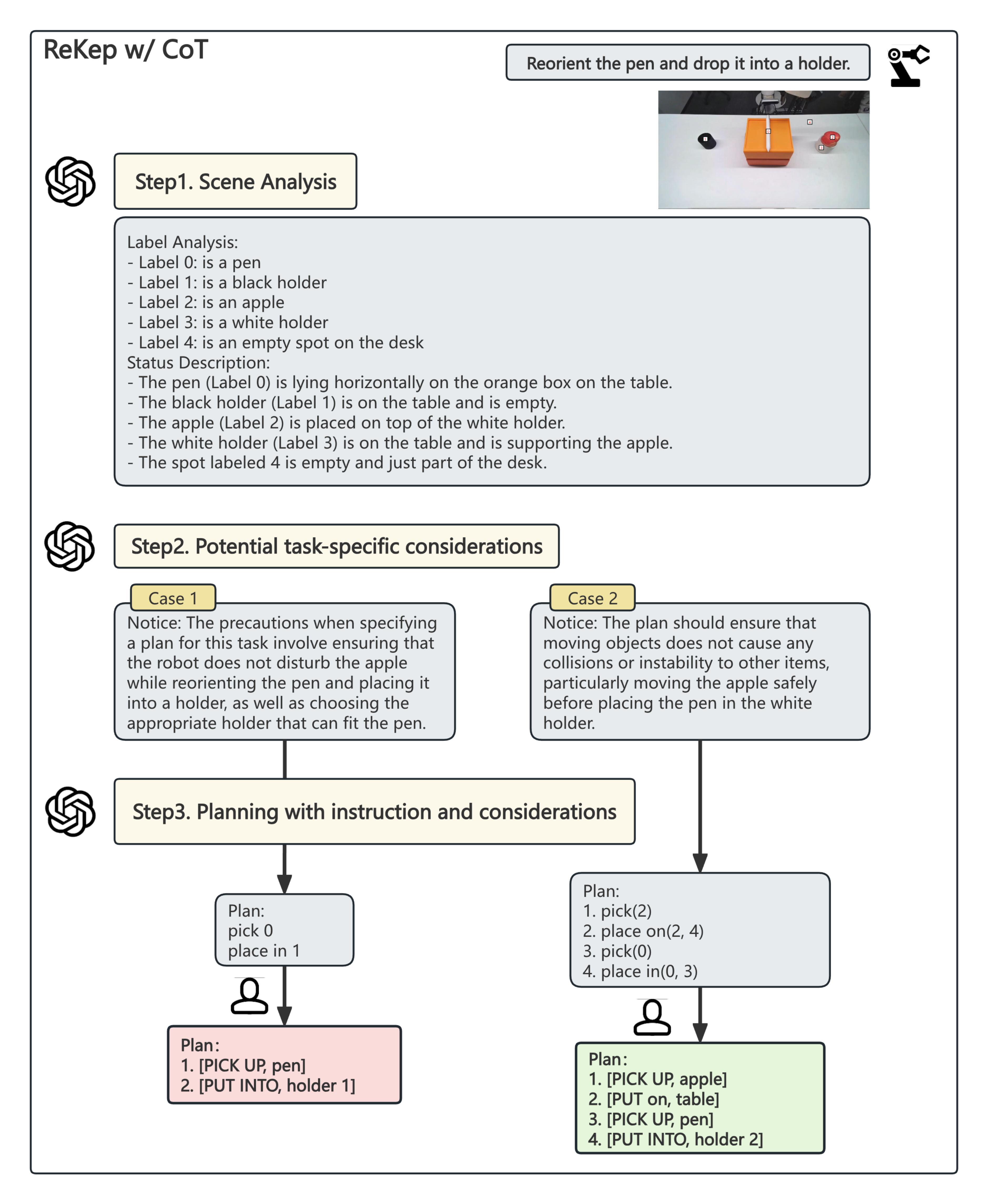}
    \caption{\textbf{Example of ReKep w/ CoT.}}
    \label{rekepwcotcase}
\end{figure*}

An example of ReKep w/ CoT planning using multi-step inference is provided in Figure~\ref{rekepwcotcase}. ReKep w/ CoT first analyzes the objects and states present in the scene, then infers potential points of attention based on the analysis and task instructions together with image observations. Finally, it generates a task plan by combining the inferred attention points, the task instructions, and the images.

The figure presents two reasoning cases: in the first case, the model produces incorrect attention points, which fail to guide the planner toward a correct solution. In the second case, the model infers correct attention points, enabling it to resolve potential issues and generate a successful plan.

\subsubsection{Reflect\(^{*}\)}

\begin{figure*}
    \centering
    \includegraphics[width=0.95\textwidth]{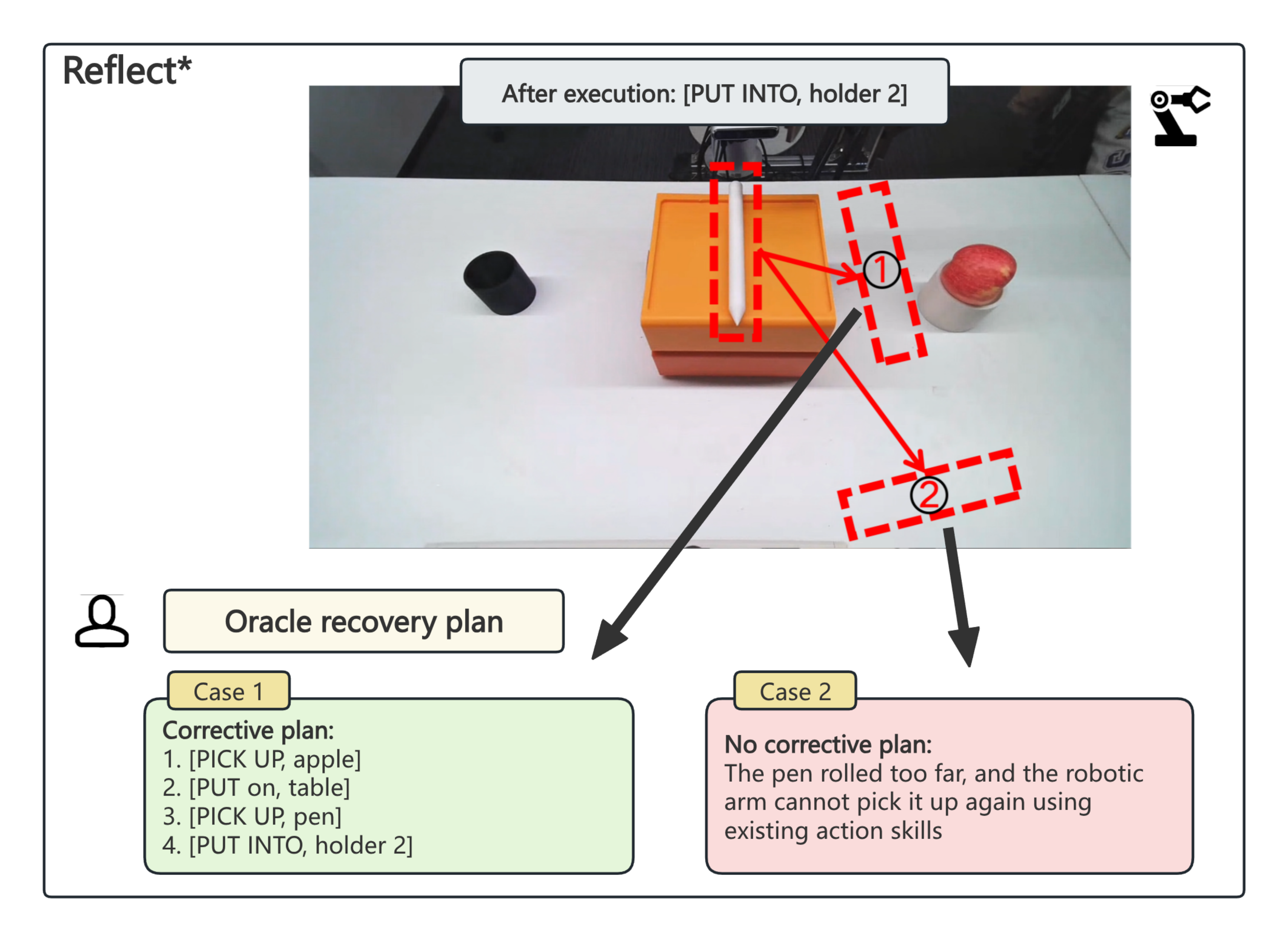}
    \caption{\textbf{Example of Reflect$^{*}$.}}
    \label{Reflectcase}
\end{figure*}

An example of Reflect\(^{*}\) is shown in Figure~\ref{Reflectcase}. After executing the initial plan  
1. [PICK UP, pen]  
2. [PUT IN, holder2]  
the second action fails in the real world. We illustrate two possible cases. When the pen slips and stops at Point~1, we can derive a feasible oracle-style recovery plan based on the available action skills, allowing Reflect$^{*}$ to successfully recover from the failure. However, when the pen slips and stops at Point~2, the distance is too large for the robot arm to generate a feasible recovery plan with existing skills, and thus Reflect$^{*}$ fails to rescue the task.

\subsubsection{EToT}

A more vivid demonstration of the EToT process is provided in the accompanying video. Please refer to the video materials and the code implementation for further details.

\subsubsection{VGM as World Model}

\begin{figure*}
    \centering
    \includegraphics[width=0.95\textwidth]{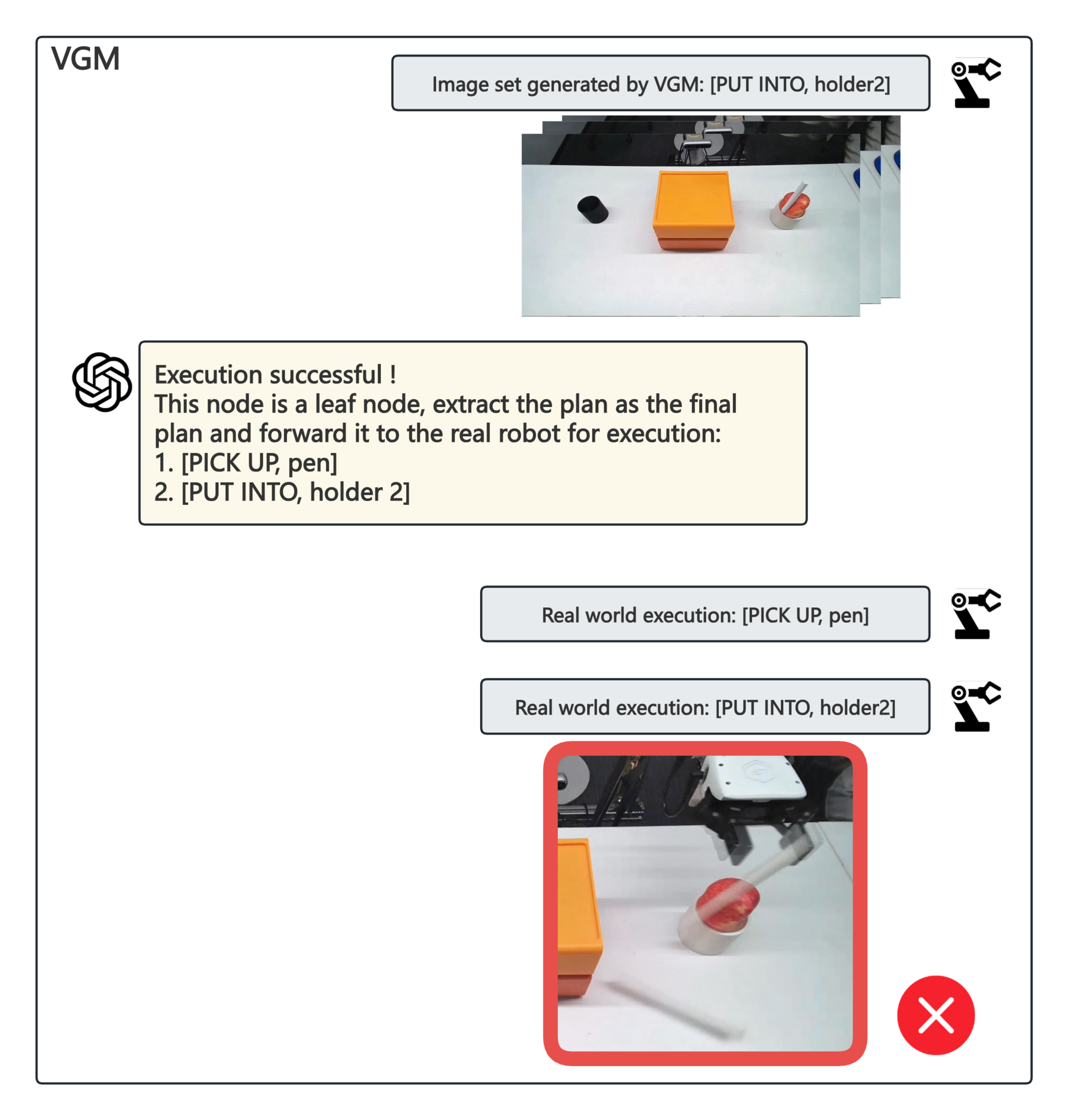}
    \caption{\textbf{Example of VGM used as a world model.}}
    \label{vgmcase}
\end{figure*}

We use the action [PUT INTO, holder2] in Task~5 as an example (Figure~\ref{vgmcase}). The prompt provided to VGM is:

\textit{“The image depicts a robotic operating environment. In the center of the desk, there is a white pen. On the left side, there is a black pen holder, and on the right side, there is a white pen holder containing an apple. The robotic arm’s gripper is positioned above the scene. Please control the robotic arm to pick up the white pen vertically, insert it into the white pen holder on the right, and then lift the gripper away from the pen.”}

The prediction generated by VGM for the action [PUT INTO, holder2] incorrectly suggests that the pen can be inserted into the pen holder even though it contains an apple, which is physically impossible. As a result, when the VLM evaluates action feasibility based on this inaccurate prediction, an incorrect plan is produced and sent to the real robot for execution, ultimately causing task failure.

\end{document}